# Objective Features Extracted from Motor Activity Time Series for Food Addiction Analysis Using Machine Learning – A Pilot Study

**Mikhail Borisenkov[1,*], Maksim Belyaev[2], Nithya Rekha Sivakumar[3,], Murugappan Murugappan[4, 5*], Andrei Velichko[2], Dmitry Korzun[6], Tatyana Tserne[7,8], Larisa Bakutova[1] and Denis Gubin[9,10,11]**

[1]Department of Molecular Immunology and Biotechnology, Institute of Physiology of Komi Science Centre of the Ural Branch of the Russian Academy of Sciences, Syktyvkar 167982, Russia
[2]Institute of Physics and Technology, Petrozavodsk State University, Petrozavodsk 185910, Russia
[3]Department of Computer Sciences, College of Computer and Information Sciences, Princess Nourah bint Abdulrahman University, Riyadh, 11671, Saudi Arabia
[4]Intelligent Signal Processing (ISP) Research Lab, Department of Electronics and Communication Engineering, Kuwait College of Science and Technology, Doha, 13133, Kuwait
[5]Department of Electronics and Communication Engineering, Faculty of Engineering, Vels Institute of Sciences, Technology, and Advanced Studies, Chennai, Tamil Nadu, India
[6]Institute of Mathematics and Information Technology, Petrozavodsk State University, Petrozavodsk 185910, Russia
[7]Pitirim Sorokin Syktyvkar State University, Syktyvkar 167000, Russia
[8]Municipal Institution of Additional Education "Center for Psychological, Pedagogical, Medical and Social Assistance", Syktyvkar 167031, Russia
[9]Laboratory for Chronobiology and Chronomedicine, Research Institute of Biomedicine and Biomedical Technologies, Medical University, Tyumen 625023, Russia
[10]Department of Biology, Medical University, Tyumen 625023, Russia
[11]Tyumen Cardiology Research Center, Tomsk National Research Medical Center, Russian Academy of Sciences, Tomsk 634009, Russia

* Correspondence: m.murugappan@kcst.edu.kw (M.M), borisenkovm@yandex.ru (M.B)

**Abstract**

Wearable sensors and IoT/IoMT platforms enable continuous, real-time monitoring, but objective digital markers for eating disorders are limited. In this study, we examined whether actimetry and machine learning (ML) could provide objective criteria for food addiction (FA) and symptom counts (SC). In 78 participants (mean age 22.1 ± 9.5 y; 73.1% women), one week of non-dominant wrist actimetry and psychometric data (YFAS, DEBQ, ZSDS) were collected. The time series were segmented into daytime activity and nighttime rest, and statistical and entropy descriptors (FuzzyEn, DistEn, SVDEn, PermEn, PhaseEn; 256 features) were calculated. The mean Matthews correlation coefficient (MCC) was used as the primary metric in a K-nearest neighbors' pipeline with five-fold stratified cross-validation (one hundred repetitions; 500 evaluations); SHAP was used to assist in interpretation. For binary FA, activity-segment features performed best (MCC = 0.78 ± 0.02; Accuracy ≈ 95.3% ± 0.5; Sensitivity ≈ 0.77 ± 0.03; Specificity ≈ 0.98 ± 0.004), exceeding OaS (Objective and Subjective Features) (MCC = 0.69 ± 0.03) and rest-only (MCC = 0.50 ± 0.03). For SC (four classes), OaS slightly surpassed actimetry (MCC = 0.40 ± 0.01 vs 0.38 ± 0.01; Accuracy ≈ 58.1% vs 56.9%). Emotional and restrained eating were correlated with actimetric features. As a result of these findings, it is possible to use wrist-worn actimetry as a digital biomarker of FA that complements questionnaires, aligning with Sensors' focus on wearable/IoT systems and supporting paths to standardized, privacy-preserving clinical translation.

**Keywords:** food addiction; actimetry; machine learning; eating behavior; depression

## 1. Introduction

Today, digital health technologies are ubiquitous, with smartphones, smartwatches, fitness bands, and textile-integrated sensors capturing physiological and behavioral signals continuously, in real-time, in everyday settings [1–4]. Current multimodal wearables incorporate on-device sensing, AI algorithms, and machine learning to transform raw streams into actionable digital biomarkers, while IoT/IoMT architecture provide wireless telemetry, cloud integration, as well as remote access to patients and clinicians [5,6]. The active (task-based) and passive (background) monitoring paradigms enhance ecological validity and support longitudinal risk stratification and intervention personalization [7]. Performance calibration and cross-device standardization remain open challenges, along with privacy, security, and user-centric adoption considerations for clinical translation at scale [3,8]. As a result, objective movement-based sensing (actimetry) offers a particularly scalable window into behavior-relevant physiology, motivating the present study. Obesity is considered a "disease of civilization", with its growth rates recently acquiring an epidemic character [9]. There is a steady increase in the number of individuals with overweight and obesity in countries with different levels of economic development and among various social and age groups [10]. Obesity is a multifactorial disease, with significant contributions from genetic, environmental, and social factors [11]. Disordered eating behaviors are also a notable risk factor for obesity [12]. Obesity is associated with challenges in human resilience to stress and mental health [13]. These challenges can be addressed through advances in Machine Learning (ML) and smart sensor technology within the Internet of Things (IoT) [14].

Most publications in recent years have focused on applying the method to identify substance [15–20] and Internet [21–25] addiction. In nutricitology, the method has been employed to evaluate the functional properties of food products [26,27,19,28–31], including component compos [26,28], caloric content [27], antioxidant activity [30], and dietary characteristics [31]. Only three publications [32–34] address issues related to food addiction. AI has been applied to assess the risk of developing food addiction associated with fast food [34] and highly processed food consumption [33]. To our knowledge, this is the first study to explore the potential of AI in identifying food addiction based on behavioral features.

Currently, the DSM-V (Diagnostic and Statistical Manual of Mental Disorders, Fifth Edition) identifies three primary eating disorders: anorexia nervosa, bulimia nervosa, and binge eating disorder (BED) [35]. The latest edition also proposes the inclusion of food addiction (FA) as a distinct disorder. The Yale Food Addiction Scale (YFAS) [36,37] and its modified version (YFAS 2.0) [38] are used to assess FA. The YFAS measures addiction-like eating of palatable foods based on the seven diagnostic criteria for substance dependence in the fourth revision of the Diagnostic and Statistical Manual of Mental Disorders (DSM-IV) [39]. Numerous studies conducted in both general and clinical populations have noted a connection between FA and BMI [37] and between FA and depression [40]. Experimental studies have identified objective neuroimaging correlates of FA [41].

However, some authors highlight difficulties in interpreting FA test results. Several studies have found no significant association between FA and BMI [41]. In some cases, FA symptoms are identified in individuals without signs of weight disorders [42], as well as in underweight individuals [43]. It has been suggested [41] that YFAS can identify disordered eating behaviors before the onset of morphological signs of obesity. However, longitudinal studies are needed to confirm this hypothesis.

FA has been shown to have a close relationship with anorexia nervosa, bulimia nervosa, and BED [44], and there is also considerable overlap between the criteria for FA and BED [45], complicating their differential diagnosis. Consequently, some experts question the necessity of recognizing FA as a separate eating disorder [46]. Other researchers [47] argue that further development of the FA concept requires improving the differential diagnosis system for eating disorders based on longitudinal study results. The diagnosis of eating disorders such as FA is often limited to questionnaire-based methods, which significantly reduce diagnostic accuracy [48]. Introducing simple and objective assessment methods for these disorders represents a crucial step in developing the FA concept. Direct and objective evaluation of the brain centers responsible for eating behavior requires costly equipment like functional magnetic resonance imaging, making it unsuitable for long-term studies. Instead, chronobiology methods, such as continuous monitoring of physiological indicators (body temperature and motor activity), are more effective, following the concept of human bionic suite composed of smart IoT devices and sensors [14,49]. Although these methods assess parameters not directly related to eating behavior, they have already been applied in this field.

The actimetry method has been extensively used to study the 24-h behavioral rhythms of patients with eating disorders, such as BED [50,51]. Literature suggests that dysfunction of the circadian system plays a key role in the etiology of BED [52]. Given the strong connections between the circadian system and eating behavior [53,54], it is expected that other forms of eating disorders are also associated with circadian system dysfunction. In our previous study, we identified a direct link between the number of FA symptoms, the mean level (MESOR), and fragmentation (intra-daily variability) of circadian motor activity [55]. These indicators could potentially differentiate between FA and BED diagnoses [55].

Currently, actimetry is mainly used in chronobiology and sleep medicine, including detecting sleep disorders. For example, early-stage Alzheimer's disease shows increased fragmentation and decreased stability in daily activity-rest rhythms [56]. Recent advancements include applying ML to automate actimetry data analysis in sleep medicine and chronobiology [57–59]. ML methods have been used for assessing driver drowsiness [58], detecting chronic insomnia [59], and analyzing rest-activity rhythms [57]. However, literature lacks ML applications for assessing eating behavior based on actimetry data. The current subjective FA assessment methods lack sensitivity. There is a need to develop objective diagnostic criteria for FA using ML for automated data collection and analysis.

This study aims to test the hypothesis that analyzing motor activity time series with an entropy-based ML algorithm can provide new criteria for FA assessment. Our team has experience with effective features, including entropy-based ones, for EEG signal analysis [60], and has developed entropy features based on NNetEn [61,62]. Entropy-based ML algorithms offer automated analysis of complex motor activity time series, revealing patterns and correlations not accessible through traditional methods [63]. This approach provides objective criteria for FA assessment and improves diagnostic accuracy by enhancing the understanding of eating behavior and neurophysiological processes.

Integrating wearable IoT devices and ML sensors further enhances this approach by enabling real-time monitoring, digital assistance, and personalized feedback, which aids in managing eating behaviors through tailored interventions. This combination of advanced ML algorithms and IoT technology offers a dynamic and comprehensive method for diagnosing and treating food addiction, ultimately contributing to improved mental health outcomes.

## 2. Materials and Methods

### *2.1. Ethics Approval Statement*

The study adhered to the tenets of the Declaration of Helsinki and research program was approved by the Ethics Committee of the Institute of Physiology of the Komi Scientific Center of the Ural Branch of the Russian Academy of Sciences (Protocol # 3, 26 March 2019). Privacy rights of human subjects have been observed. Each participant signed an informed consent form for experimentation with human subjects.

### *2.2. Study Design*

This publication presents the results of the third phase of a study focused on investigating external and internal factors associated with FA. In the first phase, climatic, socio-demographic, anthropometric, and physiological factors associated with FA were examined. In the second phase, we analyzed the relationship between metrics characterizing the daily rhythm of motor activity and FA symptoms to identify the most significant circadian rhythm indicators suitable for the differential diagnosis of FA. The results of the first and second phases were published in previous works [40,55,64]. This study used data from questionnaires conducted in the first phase and actimetry data collected from participants in the second phase.

A schematic overview of the study workflow is presented in Figure 1, with detailed steps described in the Methods and Results sections.

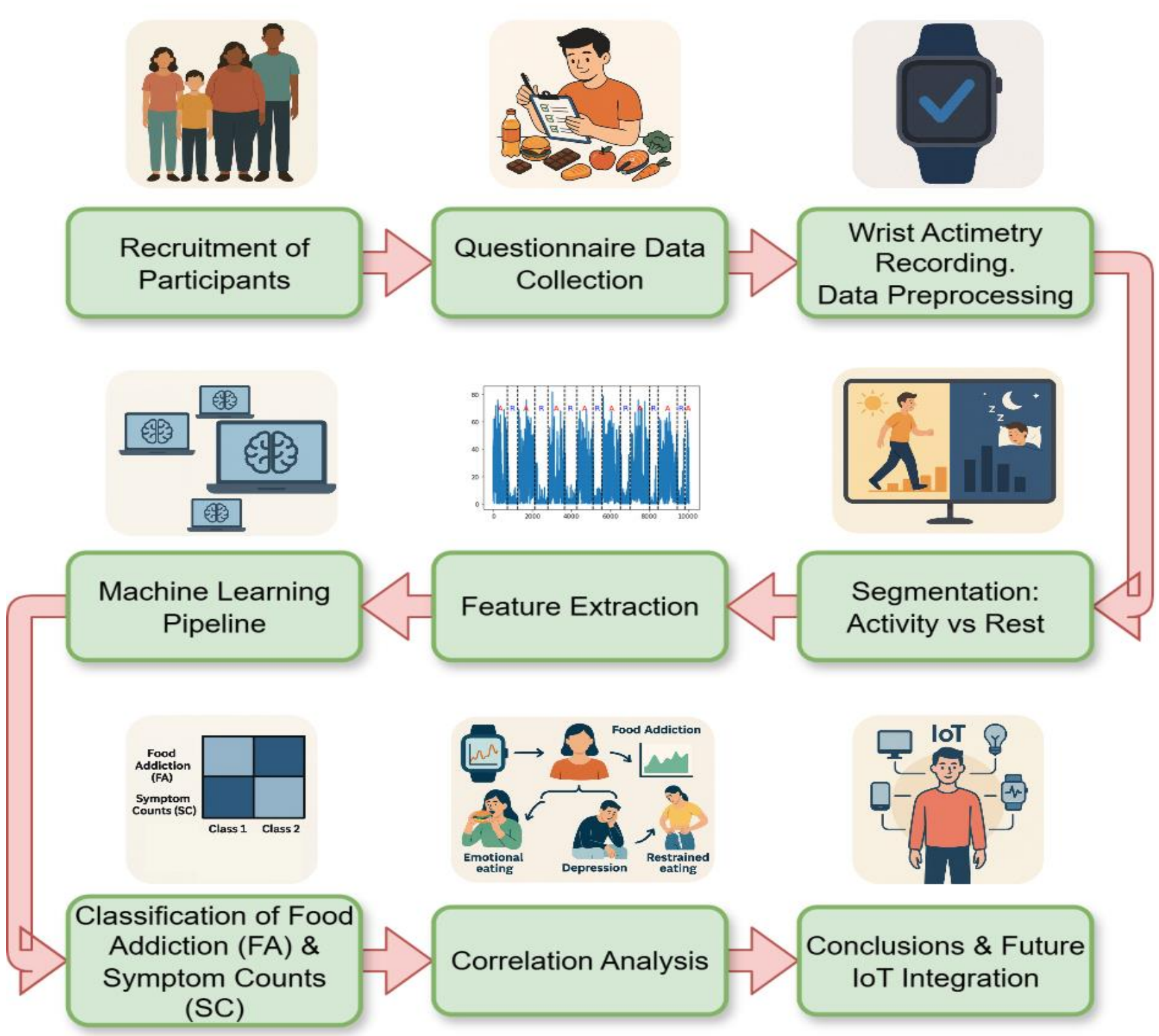

**Figure 1.** Overview of the study workflow for assessing Food Addiction (FA) using actimetric and psychophysiological data

### 2.3. *Study Participants*

The characteristics of the study participants are detailed in our previous article [55]. Briefly, data were collected from March 2019 to March 2020 in Syktyvkar and Tyumen (Lat./Long.: 61.7/50.9 and 57.2/65.5, respectively). A total of 78 participants were examined (average age: 22.1 ± 9.5, range: 18-62 yrs, women: 73.1%). Exclusion criteria included night/shift work and sleep disorders.

### 2.4. *Instruments*

Each participant provided personal data, including residence/study place, gender, age, height, and weight, and completed the YFAS questionnaire [36] and the Dutch Eating Behavior Questionnaire (DEBQ) [65]. They wore a Daqtometer v2.4 actimeter (Daqtix, Germany) on their non-dominant wrist for a week to record motor activity levels. Height and weight were used to calculate body mass index (BMI), with percentiles adjusted for gender and age using growth charts [66]. BMI categories were defined as (1) underweight, (2) normal weight, (3) overweight, and (4) obese, with categories 3 and 4 combined into a group for overweight and obesity (ov/ob).

#### 2.4.1. Yale Food Addiction Scale

The YFAS [36] was used to identify food addictions (e.g., high-fat, high-sugar, and processed foods). The Russian-translated version (YFAS-Rus) was used, with its validity confirmed by strong internal consistency and a significant correlation with the emotional eating subscale of the DEBQ [40], similar to earlier studies [37,67]. The scale has 25 items and includes 7 diagnostic criteria for substance dependence and eating behavior impairments [36]. Results are given as (a) the number of FA symptoms (SC), ranging from 0 to 7, and (b) a dichotomous measure where three or more symptoms and significant impairments indicate FA. Cronbach's alpha for this sample was 0.87.

2.4.2. Dutch Eating Behavior Questionnaire

The DEBQ [65] was used to assess eating behavior and translated into Russian (DEBQ-Rus). As shown previously [40], DEBQ-Rus has satisfactory internal consistency. The test has 33 questions divided into three subscales: restrained (DEBQrestr, 10 items), external (DEBQextern, 10 items), and emotional (DEBQemo, 13 items). A Likert scale from 1 (never) to 5 (very often) is used. Scores are averaged for each subscale. DEBQ categories (DEBQc) were assigned based on whether scores on the subscales exceeded the sample average thresholds (DEBQrestrc: 2.20, DEBQexternc: 2.93, DEBQemoc: 1.96). Cronbach's $\alpha$ value was 0.88 for DEBQ, 0.92 for DEBQrestr, 0.52 for DEBQextern, and 0.94 for DEBQemo.

2.4.3. Zung Self-Rating Depression Scale

The ZSDS which consists of 20 items was used to assess the level of depression [68]. Raw scores were transformed into ZSDS indices (ZSDSIs) as described earlier [69,70]. The ZSDSI varying from 25 to 100 scores was used as quantitative measure of the depression. In addition, ZSDSI categories (ZSDSIc; threshold value – 60 scores) were used as qualitative measure of depression. Cronbach's $\alpha$ for this sample was 0.857.

2.4.4. Wrist Actimetry

Each participant wore a Daqtometer v2.4 actimeter (Daqtix, Germany) continuously on their non-dominant wrist for one week. Activity was measured at 1 Hz, with values summed per min and expressed in arbitrary units (a.u.). The actimeter recorded dynamic (movement) and static (positional change) acceleration using a two-axis accelerometer. To quantify activity, values for each axis (xi and yi) were read every second. The linear difference between consecutive readings was summed for each 1-min interval (bin). This value was stored for each cell and computed as follows:

$$bin = \sum_{i=1}^{60} (x_i - x_{i-1}) + (y_i - y_{i-1}) \quad (1)$$

*2.5. Dataset*

The dataset under investigation contains data from 78 participants, including 13 anthropometric, psycho-emotional, and behavioral features for each individual, collected over a period of 6 to 7 d. The dataset contained actimetric data and statistical features (original dataset). The statistical features in original dataset are presented in Tables 1 and 2.

**Table 1.** General characteristics of quantitative variables

| # | Variables | Min | Max | M | SD | Missing data |
|---|---|---|---|---|---|---|
| 1 | Age | 18 | 62 | 22.14 | 9.47 | 0 |
| 2 | BMI% | 5 | 97 | 47.75 | 25.53 | 3 |
| 3 | ZSDSI | 28 | 90 | 47.27 | 12.36 | 0 |
| 4 | DEBQrestr | 1 | 4.8 | 2.20 | 1.00 | 0 |
| 5 | DEBQextern | 1.3 | 4 | 2.93 | 0.58 | 0 |
| 6 | DEBQemo | 1 | 4.2 | 1.96 | 0.85 | 0 |

BMI%: BMI percentiles, ZSDSI: Total score on the Zung Self-Rating Depression Scale, DEBQrestr: Total score on the DEBQ restrained eating subscale, DEBQextern: Total score on the DEBQ external eating subscale, DEBQemo: Total score on the DEBQ emotional eating subscale.

**Table 2.** General characteristics of qualitative variables

| # | Variables | Categories | Codes | N | % | Missing data |
|---|---|---|---|---|---|---|
| 1 | Sex | Female | 1 | 57 | 73.1 | 0 |
| | | Male | 2 | 21 | 26.9 | |
| 2 | BMIc | Underweight | 1 | 5 | 6.7 | 3 |
| | | Normal weight | 2 | 59 | 78.7 | |
| | | Overweight | 3 | 7 | 9.3 | |
| | | Obesity | 4 | 4 | 5.3 | |
| 3 | ov/ob | No | 0 | 64 | 85.3 | 3 |
| | | Yes | 1 | 11 | 14.7 | |
| 4 | ZSDSIc | No | 0 | 66 | 84.6 | 0 |
| | | Yes | 1 | 12 | 15.4 | |
| 5 | DEBQrestrc | No | 0 | 50 | 64.1 | 0 |
| | | Yes | 1 | 28 | 35.9 | |
| 6 | DEBQexternc | No | 0 | 31 | 39.7 | 0 |
| | | Yes | 1 | 47 | 60.3 | |
| 7 | DEBQemoc | No | 0 | 45 | 57.7 | 0 |
| | | Yes | 1 | 33 | 42.3 | |

BMIc: BMI categories, ov/ob: prevalence of overweight/obesity, ZSDSIc: prevalence of depression, DEBQrestrc: prevalence of restrained eating behavior, DEBQexternc: prevalence of external eating behavior, DEBQemoc: prevalence of emotional eating behavior.

The first target variable is FA, where 10 subjects were diagnosed with FA=1 (class 1), and 68 patients had no FA detected, FA=0 (class 0). The second target variable is SC. Four categories were identified based on the number of confirmed symptoms:

Class 1 (0-1 symptoms), number of records: 33;
Class 2 (2 symptoms), number of records: 16;
Class 3 (3 symptoms), number of records: 15;
Class 4 (4-7 symptoms), number of records: 14.

Since the number of elements in each class varies, classification metrics designed for imbalanced datasets were used (see section 2.8).

*2.6. Segmentation of Actimetric Data*

To extract features from actimetric data, the original time series were segmented into activity and rest periods for separate analysis. The process included:

Cleaning the data of empty values and removing inactivity periods longer than 12 h.

Calculating a moving average curve with a 1-h window.

Segmenting the time series using a change point detection algorithm based on the moving average.

Merging adjacent activity and rest segments with a threshold method.

Combining rest periods shorter than 3 h with adjacent activity periods.

Combining activity periods shorter than 4 h with adjacent rest periods.

To reduce data volatility, the actimetric curve was averaged. Segmentation used the ruptures library [71] and kernel change point detection algorithms [72,73], resulting in segmentations that could exceed the number of activity and rest periods. Change points identified transitions between different types of activity or rest phases.

The threshold method, with a global threshold set at three-quarters of the median of the actimetric curve, was used to classify segments. Segments above the threshold were classified as activity, while those below were classified as rest. Adjacent segments were merged according to specified criteria.

*2.7. Calculation of Actimetric Features*

For each participant, actimetric features were divided into two groups: activity segments (Group 1) and rest segments (Group 2) (Figure 2). Statistical methods and entropy calculation methods were employed to compute these features.

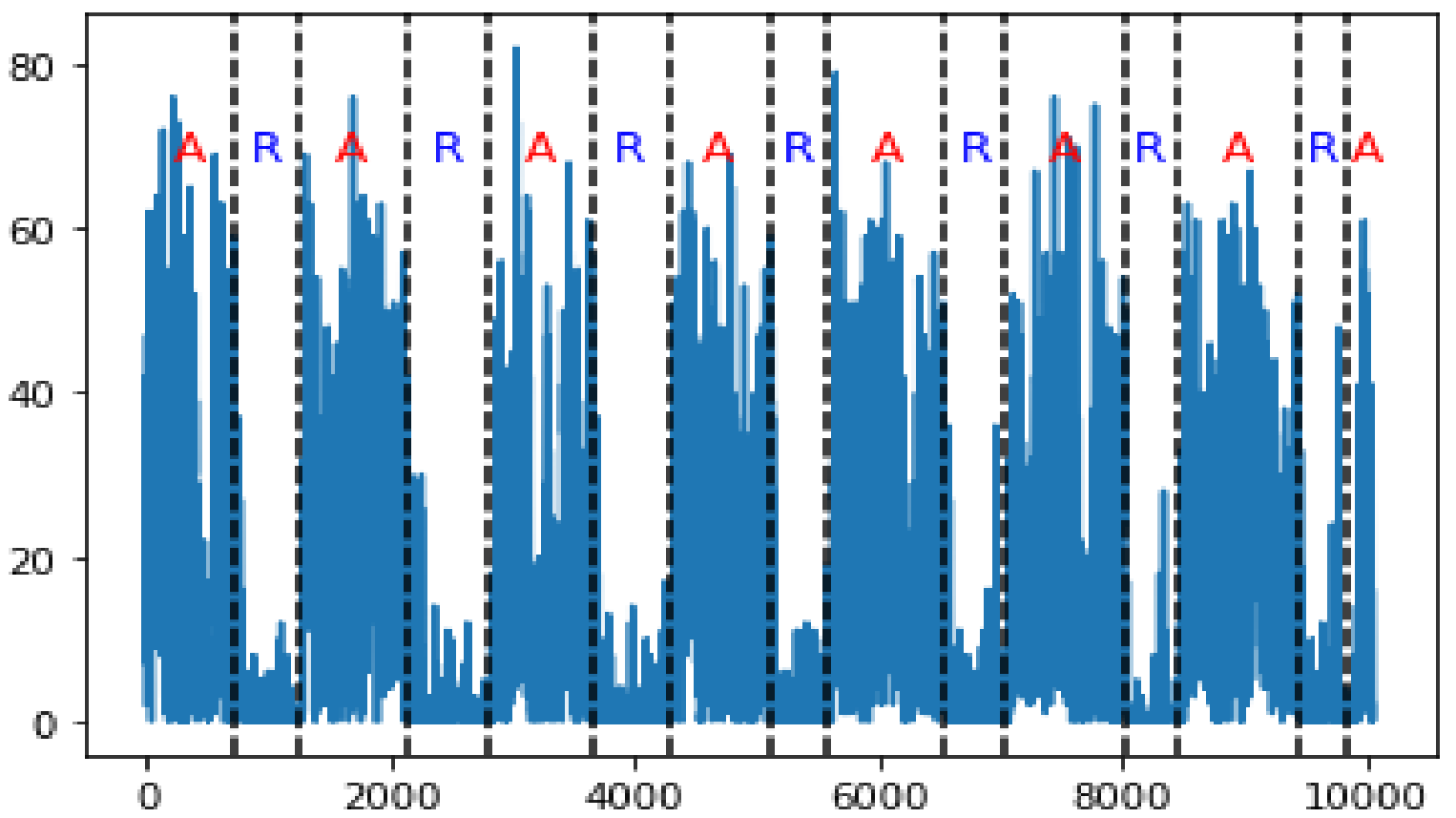

**Figure 2.** An example of actimetric data segmented into activity (A) and rest (R) segments. From a physiological point of view, the A-segment coincides with the period of daytime activity, the R-segment – with the period of nighttime sleep

The statistical methods used included the mean, maximum, minimum, range (the difference between the maximum and minimum), standard deviation, variance, coefficient of variation, and percentile values (1st, 5th, 25th, 50th, 75th, 95th, 99th percentiles). The entropy calculation methods included fuzzy entropy (FuzzyEn) [74], distribution entropy (DistEn) [75], singular value decomposition entropy (SVDEn) [76], permutation entropy (PermEn) [77], and phase entropy (PhaseEn) [78]. Each entropy calculation method was applied using ten different sets of parameters.

Let's examine the steps for calculating feature values using the activity segments (Group 1) as an example:

All activity segments from the actimetric data were selected (A1, A2, A3… AN), as shown in Figure 2. The A segments are time series of varying lengths.

For each time series segment, a statistical or entropy feature was calculated, resulting in another time series composed of these metrics. For example, TS = FuzzyEn(A1), FuzzyEn(A2), FuzzyEn(A3) … FuzzyEn(AN). Thus, a set of TS values was formed for each method and segment for each participant.

The obtained TS values for each method were aggregated using the mean (Mean(TS)) and standard deviation (Std(TS)) functions.

A similar procedure was conducted for the rest of the segments (Group 2). Consequently, each statistical and entropy method produced four features: mean and standard deviation for the activity segments, as well as mean and standard deviation for the rest segments. It is also important to note that various parameter sets were used for the entropy calculation methods to accurately characterize the dynamics of changes in participants' activity. Using this algorithm, 256 actimetric features were calculated for each record in the dataset.

### 2.8. *Machine Learning Methods*

For classifying FA and SC based on the extracted features, a ML model was implemented using the scikit-learn library in Python. The model was structured as a data processing pipeline consisting of six main stages: feature selection, handling missing values, feature scaling, classification using the chosen algorithm, cross-validation, and classification accuracy assessment.

#### 2.8.1. Feature Selection

The primary goal of this study was to identify the most significant actimetric features and their combinations for classifying FA and SC. The features were categorized into three groups: actimetric features from activity segments, actimetric features from rest segments, and subjective features. A sequential search was conducted across various combinations of these feature groups to determine their impact on the classification results.

#### 2.8.2. Handling Missing Values

The original dataset contained missing values in the subjective features (see Tables 1 and 2). These gaps were filled using the mean values for each feature to ensure consistency across the dataset.

#### 2.8.3. Feature Scaling

Scaling of feature values was necessary when the model included more than one feature. The min-max scaling method was applied, normalizing feature values to a range between 0 and 1 to ensure comparability.

#### 2.8.4. Classification Algorithm

The K-Nearest Neighbors (KNN) algorithm [79] was selected for classification due to its simplicity and efficiency. KNN classifies data points based on their proximity to other samples in the training set.

#### 2.8.5. Cross-Validation

Repeated stratified k-fold cross-validation was used to estimate model performance. The dataset was partitioned into $K = 5$ stratified folds; in each run, the model was trained on ~80% of the data and tested on the remaining ~20%. The process was repeated N = 100 times, each with a different random fold assignment (total $K \cdot N = 500$ train–test evaluations). As a primary criterion for selecting the most informative features and their combinations, Matthew's correlation coefficients (MCCs) [80] were computed for each evaluation. Across repetitions, 95 percent confidence intervals were obtained. Hyperparameters were also assessed for model accuracy, and for each feature set, the configuration yielding the highest mean MCC was chosen.

#### 2.8.6. Classification Accuracy Assessment

Classification performance was primarily assessed by the mean Matthews correlation coefficient (MCC). In addition, we report mean Accuracy, Sensitivity, Specificity, and F1-score, together with the confusion matrix averaged across all cross-validation folds and repetitions.

In binary classification, there are four possible outcomes: correct predictions of positive samples (True Positive, TP) and negative samples (True Negative, TN), as well as incorrect predictions of negative samples (False Positive, FP) and positive samples (False Negative, FN). The MCC is calculated using the following formula:

$$MCC = \frac{TP \cdot TN - FP \cdot FN}{\sqrt{(\mathrm{TP+FP}) \cdot (\mathrm{TP+FN}) \cdot (\mathrm{TN+FP}) \cdot (\mathrm{TN+FN})}} \quad (2)$$

This metric is particularly valuable for evaluating model accuracy with imbalanced datasets because it considers all four outcomes, including True Negatives, unlike the F1-score. MCC was calculated using standard libraries from scikit-learn for both binary and multiclass tasks, providing a comprehensive assessment of model performance.

2.8.7. Single-Feature and Feature-Combination Analyses

Alongside the mean MCC for individual features, their statistical significance was assessed using the Kruskal–Wallis test, with p-values and effect sizes reported. Effect size was quantified by eta-squared as defined in equation (3);

$$\eta^2 = \frac{H-k+1}{n-k}, \quad (3)$$

In that expression, $H$ denotes the Kruskal–Wallis H statistic, $k$ the number of groups (classes), and $n$ the total number of observations (n = 78). To quantify the contribution of individual features within multivariate models for FA classification, we employed SHAP (Shapley Additive Explanations). To enhance reliability, SHAP values for each feature and each sample were averaged across $N = 100$ repetitions of the cross-validation procedure. Results were summarized using beeswarm plots.

## 3. Numerical Results

### *3.1. Segmentation Results of Actimetric Data*

Figure 3 shows examples of the segmentation of actimetric data from the studied dataset, illustrating various levels of FA and the number of confirmed SC. The visualization is presented in the form of three columns, each corresponding to different levels of FA and SC:

FA=0, SC=1: This chart shows data from a participant with no food addiction and minimal symptoms. It reveals regular activity and rest cycles with clear segmentation, indicating a balanced daily routine.

FA=0, SC=6: This chart depicts a participant without food addiction but with many symptoms. Increased fragmentation and instability in activity rhythm suggest possible circadian disturbances despite no FA.

FA=1, SC=3: This example features a participant with food addiction and moderate symptoms. The chart shows significant activity fluctuations and prolonged periods, reflecting impulsive or irregular behavior typical of FA.

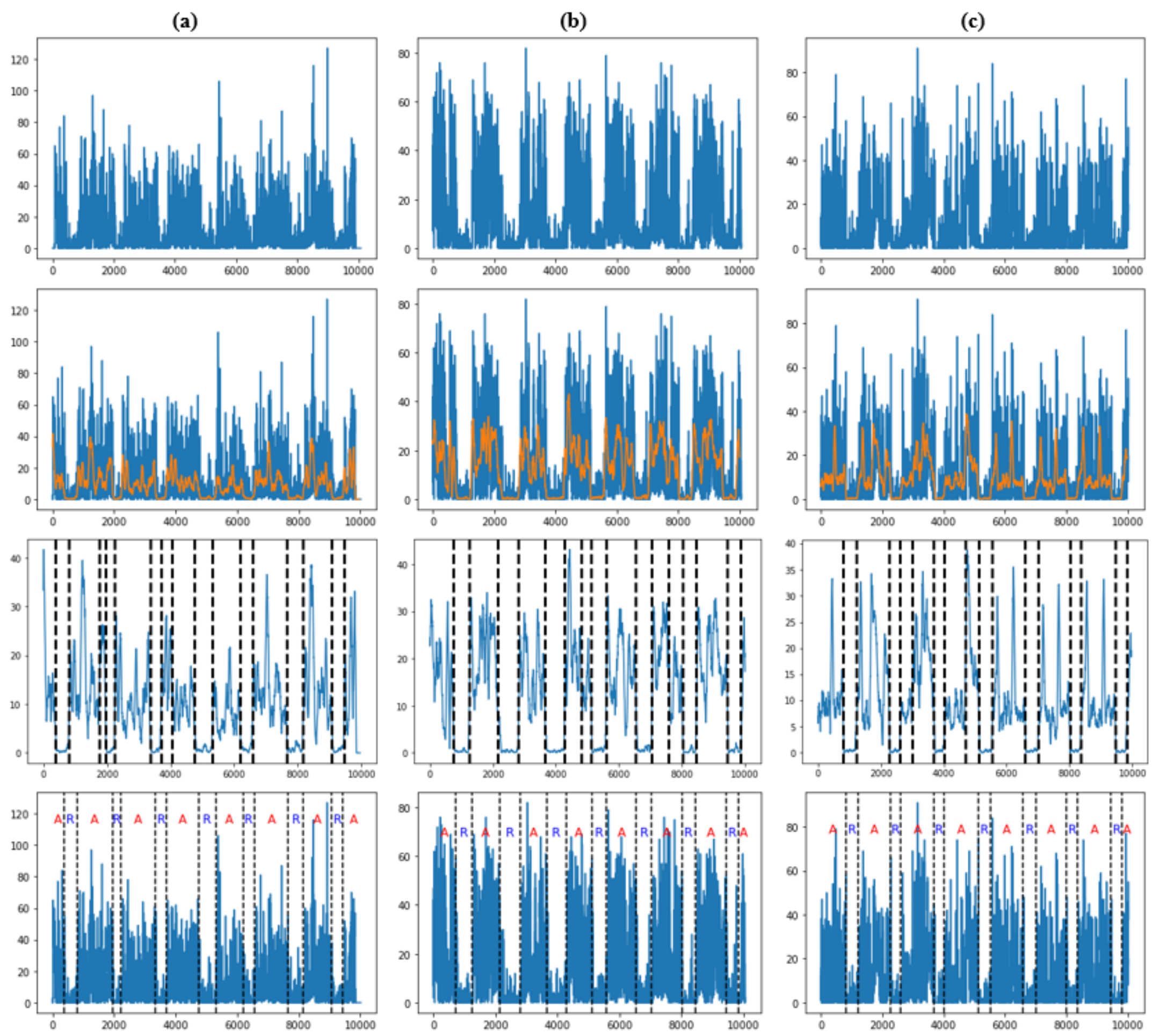

**Figure 3.** Examples of actimetry data segmentation from the study dataset with varying levels of food addiction (FA) and symptom counts (SC). The figure contains four rows of plots arranged in three columns: (a) FA = 0, SC = 1; (b) FA = 0, SC = 6; (c) FA = 1, SC = 3. The top row shows the original actimetry time series data; the second row presents the same data overlaid with a moving average curve (orange line); the third row illustrates segmentation of the moving average curve using the kernel change point detection method; and the bottom row displays the final segmentation of the original data after merging segments, with activity (A) and rest (R) periods indicated. FA – food addiction; SC – symptom counts of food addiction.

Data segmentation involved several steps: cleaning gaps (e.g., for water procedures), removing erroneous inactivity periods over 12 h, and using a change point detection algorithm to identify state changes. Adjacent activity and rest segments were then merged, smoothed, and stabilized. A median-based threshold method combined segments effectively, distinguishing phases of activity and rest even in imbalanced datasets, as described in section 2.8.

Overall, the charts demonstrate the proposed method's effectiveness in distinguishing and visualizing different levels of FA and related SC, facilitating detailed analysis of actimetric data in eating behavior studies.

### *3.2. FA Classification Results*

#### 3.2.1. Classification Results of Individual Features

After segmenting the data and calculating the actimetric features, the entire dataset was analyzed using a classification algorithm to assess the impact of each individual feature. Tables 3, 4, and 5 present the six most important features from the original dataset of subjective features (see Tables 1 and 2) and actimetric features, divided into activity and rest segments.

**Table 3.** The most significant subjective features for determining FA from the original dataset

| Feature Name | MCC | p-value | Effect size |
|---|---|---|---|
| DEBQrestr | 0.31 ± 0.03 | 0.0211 | 0.06 |
| ZSDSIc | 0.27 ± 0.03 | 0.0012 | 0.12 |
| ZSDSI | 0.23 ± 0.03 | 0.0002 | 0.17 |
| DEBQextern | 0.21 ± 0.03 | 0.0067 | 0.08 |
| DEBQexternc | 0.20 ± 0.02 | 0.0408 | 0.04 |
| DEBQemo | 0.14 ± 0.03 | 0.0043 | 0.09 |

The analysis used variables listed in Tables 1 and 2; MCC: Matthew's correlation coefficient, other abbreviations as in Tables 1 and 2.

**Table 4.** The most significant actimetric features for determining FA based on activity segments

| Aggregation Function | Feature | MCC | p-value | Effect size |
|---|---|---|---|---|
| Standard Deviation | Maximum Value | 0.58 ± 0.03 | 0.7994 | -0.01 |
| Standard Deviation | FuzzyEn | 0.56 ± 0.03 | 0.1471 | 0.01 |
| Mean Value | SVDEn | 0.39 ± 0.03 | 0.3778 | -0.003 |
| Mean Value | 99th Percentile | 0.29 ± 0.03 | 0.0617 | 0.03 |
| Standard Deviation | DistEn | 0.29 ± 0.03 | 0.7879 | -0.01 |
| Mean Value | Standard Deviation | 0.29 ± 0.03 | 0.0136 | 0.07 |

FuzzyEn: fuzzy entropy, SVDEn: singular value decomposition entropy, DistEn: distribution entropy, MCC: Matthew's correlation coefficient.

**Table 5.** The most significant actimetric features for determining FA based on rest segments

| Aggregation Function | Feature | MCC | p-value | Effect size |
|---|---|---|---|---|
| Mean Value | FuzzyEn | 0.44 ± 0.03 | 0.6755 | -0.01 |
| Standard Deviation | Standard Deviation | 0.31 ± 0.03 | 0.1513 | 0.01 |
| Standard Deviation | DistEn | 0.31 ± 0.03 | 0.8694 | -0.01 |
| Standard Deviation | FuzzyEn | 0.30 ± 0.03 | 0.3024 | 0.001 |
| Standard Deviation | 75th Percentile | 0.27 ± 0.03 | 0.9027 | -0.01 |
| Mean | 95th Percentile | 0.20 ± 0.03 | 0.7286 | -0.01 |

Abbreviations as in Table 4.

Analyzing individual features shows that actimetric features (Tables 4 and 5) have higher MCC values than features from the original dataset (Table 3). The highest MCC value of 0.58 ± 0.03 is for activity segments (standard deviation of the maximum value). Other features from activity segments also show higher MCC values, highlighting their greater importance for FA classification compared to rest segments.

Notably, the highest accuracy for activity segments corresponds to the standard deviation as the aggregation function, indicating that variation in daily activity indicators is crucial for determining FA. For rest segments, the most important feature is the mean value of FuzzyEn.

Several top actimetric features show high MCCs along with non-significant univariate p-values (Tables 4, 5). This is expected when class structure is primarily multivariate and non-monotonic: class-conditional distributions are multimodal (in our data often bimodal), so marginal location tests (e.g., Kruskal–Wallis) have low power, while an instance-based classifier such as KNN exploits cluster geometry and feature interactions. Discriminative rules, in practice, represent "two-threshold" regions (values ≤ a or ≥ b) that single-feature statistics do not capture. Multimodality was confirmed by density/violin plots, and feature–feature interactions were indicated by SHAP dependence plots. The informative signal, therefore, comes from the combination of actimetric features - notably during activity segments - which supports the use of multivariate, nonlinear models for FA assessment alongside (but not necessarily limited to) univariate significance testing.

#### 3.2.2. Classification Results of Feature Combinations

In the next stage of the work, a search was conducted for feature combinations that provide the highest classification accuracy. Table 6 presents the results for all four groups.

**Table 6.** The most significant feature combinations for determining FA

| Group | Features | MCC |
|---|---|---|
| Objective and Subjective Features | ZSDSIc, DEBQexternc, DEBQemo, BMI% | 0.69 ± 0.03 |
| Actimetric Features, Activity Segments Only | Standard Deviation of FuzzyEn, Mean Value of the 99th Percentile, Mean Value of the 5th Percentile | 0.78 ± 0.02 |
| Actimetric Features, Rest Segments Only | Mean Value of FuzzyEn, Standard Deviation of the Standard Deviation | 0.50 ± 0.03 |

Abbreviations as in Tables 1, 2 and 4.

Table 6 shows that the highest MCC value (0.78) is achieved using features from activity segments, offering better classification accuracy than the original dataset features (Tables 1 and 2). The lowest MCC value (0.50) is from features derived from rest segments, highlighting the stronger link between FA and activity segments. Using all actimetric features (for activity and res segments) does not improve MCC value. In cases with an MCC of 0.78, the confusion matrix is shown in Table 7 and has slightly more false negative errors (2.32), than false positive errors (1.37). Classification accuracy is (95.3±0.5)%, sensitivity is 0.77±0.03, specificity is 0.98±0.004, and an F1-score is 0.78±0.02.

**Table 7.** Averaged confusion matrix based on classification results (Actimetric Features, Activity Segments Only)

| | | Predicted Values | |
|---|---|---|---|
| | | **FA=0** | **FA=1** |
| Actual Values | FA=0 | 66.63 | 1.37 |
| | FA=1 | 2.32 | 7.68 |

FA: food addiction.

In the OaS model (Figure 4a) the SHAP summary, despite some scatter expected with $n = 78$, shows consistent directions for several variables. Lower ZSDSIc values (blue) tend to shift predictions toward FA (positive

SHAP), whereas higher ZSDSIc are neutral or protective. Higher DEBQemo contributes slightly positively, while higher DEBQexternc contributes slightly negatively, and BMI% has only a minor, near-zero effect. This pattern matches the modest univariate MCC of subjective variables and suggests their influence is real but comparatively weak and partly mediated by interactions.

For the activity-only model (Figure 4b) the ordering by color is clearer. Lower values of the standard deviation of FuzzyEn (blue) push the model toward FA, whereas higher variability in FuzzyEn is protective. Higher extremes of activity (larger mean of the 99th percentile) and a raised low-end baseline (larger mean of the 5th percentile) both tend to increase FA probability (positive SHAP). Taken together, the combination "reduced variability in entropy across active periods + occasional very high bouts + elevated baseline" is associated with FA classification, which is consistent with the superior MCC of activity-segment features. The residual dispersion and occasional reversals in color gradients are consistent with multimodal, non-monotonic relationships that KNN can capture locally, while univariate statistics may understate them.

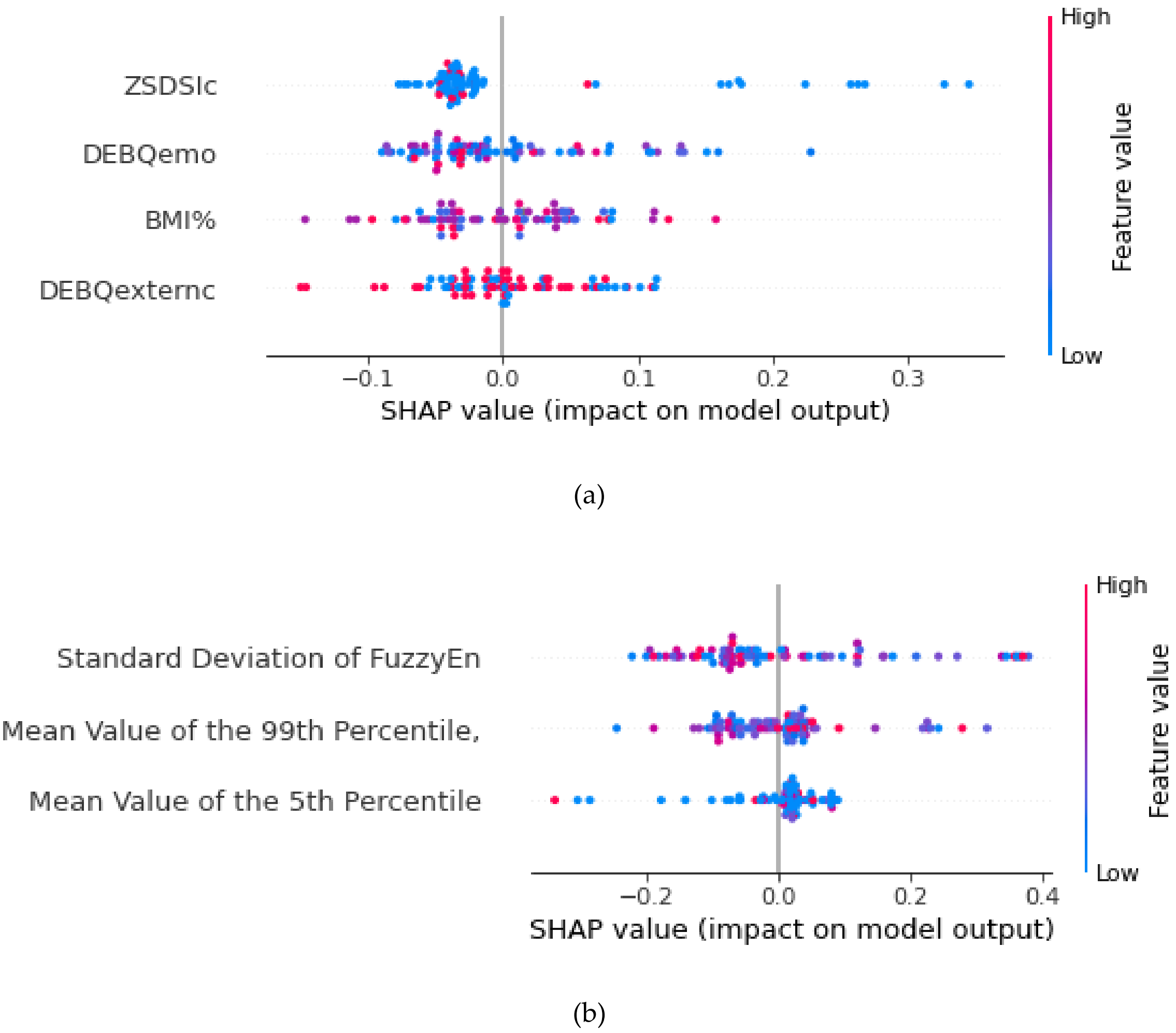

**Figure 4.** Mean SHAP summary plots for (a) the OaS model (Objective-and-Subjective features: ZSDSIc, DEBQexternc, DEBQemo, BMI%) and (b) the activity-only actimetric model (standard deviation of FuzzyEn; mean of the 99th and 5th percentiles). Each dot denotes one participant; color encodes the feature value (blue =

low, red = high); the x-axis shows the SHAP value (impact on the FA = 1 prediction). SHAP values were averaged across 100 repetitions of stratified five-fold cross-validation. In (b), lower variability of FuzzyEn and higher extremes/baseline of activity (99th/5th percentiles) shift predictions toward FA, aligning with the superior MCC of activity-segment features.

### 3.3. *SC Classification Results*

#### 3.3.1. Classification Results of Individual Features

Tables 8, 9, and 10 present the six most significant features (based on MCC values) from the original dataset (see Table 1), as well as actimetric features divided into activity and rest segments.

**Table 8.** The most significant features for determining SC from the original dataset

| Feature Name | MCC | p-value | Effect size |
|---|---|---|---|
| DEBQextern | 0.23 ± 0.01 | 0.0029 | 0.15 |
| ZSDSI | 0.23 ± 0.01 | 0.0046 | 0.14 |
| Age | 0.21 ± 0.01 | 0.0005 | 0.20 |
| ov/ob | 0.16 ± 0.01 | 0.0020 | 0.16 |
| BMIc | 0.12 ± 0.01 | 0.0014 | 0.17 |
| BMI% | 0.12 ± 0.01 | 0.0078 | 0.12 |

The analysis used variables listed in Tables 1 and 2.

**Table 9.** The most significant actimetric features for determining SC based on activity segments

| Aggregation Function | Feature | MCC | p-value | Effect size |
|---|---|---|---|---|
| Standard Deviation | 95th Percentile | 0.27 ± 0.01 | 0.0197 | 0.09 |
| Mean Value | DistEn | 0.24 ± 0.01 | 0.0386 | 0.07 |
| Mean Value | PermEn | 0.22 ± 0.01 | 0.6004 | -0.02 |
| Standard Deviation | FuzzyEn | 0.22 ± 0.01 | 0.2214 | 0.02 |
| Mean Value | Mean Value | 0.21 ± 0.01 | 0.2291 | 0.02 |
| Mean Value | 99th Percentile | 0.20 ± 0.01 | 0.0077 | 0.12 |

**Table 10.** The most significant actimetric features for determining SC based on rest segments

| Aggregation Function | Feature | MCC | p-value | Effect size |
|---|---|---|---|---|
| Mean Value | FuzzyEn | 0.27 ± 0.02 | 0.2961 | 0.01 |
| Standard Deviation | FuzzyEn | 0.25 ± 0.01 | 0.5289 | -0.01 |
| Standard Deviation | PhaseEn | 0.18 ± 0.01 | 0.6910 | -0.02 |
| Mean Value | PhaseEn | 0.17 ± 0.01 | 0.8110 | -0.03 |
| Standard Deviation | SVDEn | 0.17 ± 0.01 | 0.5475 | -0.01 |
| Mean Value | Standard Deviation | 0.15 ± 0.01 | 0.0293 | 0.08 |

Abbreviations as in Table 4.

#### 3.3.2. Classification Results of Feature Combinations

In the next stage of the work, a search was conducted for feature combinations that provide the highest classification accuracy. Table 11 presents the results for all four groups.

**Table 11.** The most significant feature combinations for determining SC

| Group | Features | MCC |
|---|---|---|
| Objective and Subjective Features | Age, BMIc, ZSDSI, DEBQextern, DEBQemoc | 0.40 ± 0.01 |
| Actimetric Features, Activity Segments Only | Standard Deviation of the 95th Percentile,<br>Mean Value of the 95th Percentile<br>Standard Deviation of PermEn | 0.38 ± 0.01 |
| Actimetric Features, Rest Segments Only | Mean Value of FuzzyEn,<br>Standard Deviation of the 95th Percentile | 0.28 ± 0.01 |

Abbreviations as in Tables 1, 2 and 4.

Table 11 shows that the highest MCC value (0.40 ± 0.01) is achieved with objective and subjective (OaS) dataset features, offering slightly better classification accuracy than the features from actimetric features for activity segments (MCC=0.38 ± 0.01). The lowest MCC value (0.28 ± 0.01) is for features from rest segments, highlighting the greater significance of activity segments in determining SC. Using all actimetric features (for activity and res segments) does not improve MCC value. The average confusion matrix for classification using activity segments is shown in Table 12.

**Table 12.** Averaged confusion matrix for SC classification using actimetric features from activity segments.

| | | Predicted Values | | | | Sensitivity | Specificity | F1-score |
|---|---|---|---|---|---|---|---|---|
| | | **SC=1** | **SC=2** | **SC=3** | **SC=4** | | | |
| Actual Values | SC=1 | **29.0** | 2.37 | 0.74 | 0.89 | 0.88±0.01 | 0.50±0.01 | 0.68±0.01 |
| | SC=2 | 3.29 | **11.65** | 0.07 | 0.99 | 0.73±0.02 | 0.88±0.01 | 0.67±0.02 |
| | SC=3 | 9.93 | 3.76 | **1.02** | 0.29 | 0.07±0.01 | 0.98±0.01 | 0.09±0.02 |
| | SC=4 | 9.57 | 1.02 | 0.69 | **2.72** | 0.20±0.02 | 0.97±0.01 | 0.26±0.02 |

SC: symptom counts of food addiction.

**Table 13.** Averaged confusion matrix for SC classification using objective and subjective features.

| | | Predicted Values | | | | Sensitivity | Specificity | F1-score |
|---|---|---|---|---|---|---|---|---|
| | | **SC=1** | **SC=2** | **SC=3** | **SC=4** | | | |
| Actual Values | SC=1 | **28.69** | 0.30 | 3.03 | 0.98 | 0.87 ± 0.01 | 0.49± 0.01 | 0.68± 0.01 |
| | SC=2 | 9.23 | **5.19** | 0.62 | 0.96 | 0.32 ± 0.02 | 0.97 ± 0.01 | 0.40 ± 0.02 |
| | SC=3 | 10.23 | 1.4 | **1.96** | 1.41 | 0.13 ± 0.02 | 0.93 ± 0.01 | 0.16 ± 0.02 |
| | SC=4 | 3.39 | 0.22 | 0.88 | **9.51** | 0.68 ± 0.02 | 0.95 ± 0.01 | 0.69 ± 0.02 |

The analysis used variables listed in Tables 1 and 2. SC: symptom counts of food addiction.

- **Findings from the confusion matrices (activity-only vs OaS models):** Across the four SC classes, both models recall the majority class SC=1 well (Sensitivity ≈0.87–0.88; F1 ≈0.68) but show low specificity for this class (≈0.49–0.50) because many non-SC1 cases are predicted as SC1. The activity-only actimetric model excels at detecting SC=2 (Sensitivity 0.73, F1 0.67) but performs poorly for higher counts (SC=3–4; Sensitivity 0.07 and 0.20, respectively). In contrast, the OaS model markedly improves detection of SC=4 (Sensitivity 0.68, F1 0.69) while being weaker on SC=2 (Sensitivity 0.32). For SC=3, both models underperform (F1 ≤ 0.16), suggesting substantial class overlap and/or scarcity.

- **Overall comparison and implications:** The OaS model achieves slightly higher mean performance (Accuracy 58.1% vs 56.9%; MCC 0.40 vs 0.38), consistent with these classwise patterns. A hybrid approach (e.g., stacking or an ordinal objective) could rebalance sensitivities across adjacent symptom-count classes using complementary error profiles-activity features favoring SC=2 and OaS features favoring SC=4. Considering the ordinal nature of SC (1–4) and the concentration of errors into neighboring classes (especially SC=1), ordinal classification or cost-sensitive training may reduce confusion between adjacent categories and improve macro-F1 without sacrificing overall accuracy.

### *3.4. Relationship Between Actimetry and Subjective Features*

To analyze the relationship between the OaS features (BMI%, ZSDSI, DEBQrestr, DEBQextern, DEBQemo) from Table I and the most significant actimetry features for determining FA or SC (Tables 4, 5, 9 and 10), Pearson's correlation analysis was used (Table 14).

**Table 14.** Results of the correlation analysis of the relationships between the studied indicators

| Aggregation Function | Feature | Subjective Features | | | | |
|---|---|---|---|---|---|---|
| | | **BMI%** | **ZSDSI** | **DEBQrestr** | **DEBQextern** | **DEBQemo** |
| | | Segments of activity | | | | |
| *SD* | Max | - | - | - | - | 0.263* |
| *SD* | FuzzyEn | - | - | - | 0.265* | - |
| *M* | SVDEn | -0.296** | - | - | - | - |
| *M* | 99th percentile | - | - | - | - | 0.303** |
| *M* | *SD* | - | - | 0.289* | - | 0.382** |
| *M* | *M* | - | - | 0.288* | - | - |
| | | Segments of rest | | | | |
| *M* | SD | - | - | 0.267* | - | 0.273* |

The table presents Pearson correlation coefficients for the features BMI%, ZSDSI, DEBQrestr, DEBQextern, DEBQemo, and the most significant actimetry features taken from Tables 4, 5, 9 and 10, segmented by activity and rest. * - $P < 0.05$, ** - $P < 0.01$; the rest of the abbreviations are as in Tables 1, 2 and 4.

Table 14 demonstrates significant correlations between certain actimetry and subjective features, highlighting the importance of some metrics for assessing behavioral and physiological parameters. Among the actimetry features, significant correlations are observed in the activity segments for DEBQemo, BMI%, DEBQrestr, and DEBQextern. For DEBQemo and DEBQrestr, the correlation is positive, with the highest values observed for the mean (M) of the standard deviation (SD). A positive correlation is also observed for DEBQextern and the SD of FuzzyEn. For BMI%, a negative correlation is observed with the M of SVDEn. In the rest segments, significant correlations were found only between DEBQemo, DEBQrestr, and the M of variance. It can be noted that the strongest linear correlations are between the actimetry features of the activity segments and the initial features.

## 4. Discussion

### 4.1. Relationship between Motor/Physical Activity and Food Addiction

This study shows that actimetry-derived characteristics of motor activity contain clinically useful information for assessing disordered eating—specifically, food addiction (FA). Leveraging AI/ML to analyse continuous sensor streams (feature engineering of statistical and entropy measures, KNN classification, and SHAP-based interpretation) allows detection of multivariate, non-monotonic patterns that are not apparent in single-feature statistics. Our findings are consistent with prior work [55] showing that individuals with FA exhibit two distinctive properties of the 24-h activity profile—elevated Midline Estimating Statistic of Rhythm (MESOR) and increased intradaily variability (IV). Earlier studies have noted similarities between FA and binge-eating disorder (BED)—greater impulsivity [39,41,81] and a tendency toward depression [39]—yet these disorders differ in their behavioural component [45]. Importantly, daily activity rhythms in BED, unlike FA, are characterised by lower MESOR [51,51] and higher interdaily stability [51]. Taken together, these observations suggest that combining actimetry with AI could support a practical algorithm for differential diagnosis of BED and FA, and potentially other eating-disorder phenotypes.

Several questionnaire-based studies [51,82,83] reported that adults with FA tend to be more sedentary than peers without FA, whereas one study [84] found a positive association between FA and activity level. Such discrepancies are likely driven by limitations of self-report (recall bias, coarse intensity scales, insensitivity to circadian structure). Objective, continuous actigraphy addresses these issues, and our results indicate that risk is linked not only to overall volume but also to the distribution and rhythmic organisation of activity—features that AI methods can extract from sensor data to yield robust, interpretable digital markers of FA.

### 4.2 Relationship Between Actimetric Features of FA and Psychoemotional Characteristics

The most significant correlations were found between actimetric features and emotional eating behavior, confirming its well-documented link to FA [39,85]. Notably, there were also significant correlations with restrained eating behavior, emphasizing its relevance to FA diagnosis [86]. We observed only one significant correlation between an actimetric feature for FA and BMI%, related to activity segments, which aligns with prior studies showing either a lack of or nonlinear relationship between FA and anthropometric characteristics [43]. Our study did not find significant correlations between FA and depression, contrary to earlier research suggesting a strong link [40]. This may indicate a need for further investigation into the complex relationship between FA and depression.

### 4.3. Analysis of FA and SC Classification Accuracy

The analysis of data from Tables 4, 5, 9 and 10 reveals that MCC values for features from activity segments are slightly higher than for rest segments, indicating a stronger connection between FA and SC and motor activity characteristics during the day. This is further evident in the optimal combinations of actimetric features (Tables 6 and 11). Combining features from activity and rest segments does not improve the MCC metric. Thus, daytime activity alone is sufficient for classifying FA and SC.

Comparing OaS (Tables 1 and 2) and actimetric features for FA (Tables 3-6) classification shows that actimetric features enable more accurate classification. For SC (Tables 8-11), the maximum MCC values are close: MCC=0.27 (actimetry) vs. MCC=0.23 (OaS features) for individual features, and MCC=0.38 (actimetry) vs. MCC=0.40 (OaS features) for optimal combinations. For FA, the difference is more pronounced: MCC=0.58 (actimetry) vs. MCC=0.31 (OaS features) for individual features, and MCC=0.78 (actimetry) vs. MCC=0.69 (OaS features) for optimal combinations. The maximum MCC values for SC are significantly lower than for FA, likely due to more classes (4 for SC vs. 2 for FA) and difficulty in classifying SC classes {2,3,4} (Tables 12 and 13).

### *4.3. Strengths and Limitations*

The proposed method accurately identifies individuals with food addiction with high specificity and sensitivity. A key advantage of the study is its inclusion of participants from two geographically distant regions. Analysis results show that while basic demographic and anthropometric features (such as age, depression level, and body mass index) have moderate significance, actimetric features related to activity segments offer more precise indicators for assessing FA symptoms. This highlights the need for further research and the integration of objective motor activity data into diagnostic algorithms for better assessment of eating disorders. However, the study has limitations, including a small sample size, gender imbalance, and a cross-sectional design that prevents conclusions about causal relationships.

## 5. Conclusion

This study demonstrated that actimetric data combined with machine learning can be used to produce an accurate model that reflects the relationship between food addiction and psychophysiological characteristics. According to our findings, the actimetric features identified could serve as objective criteria for FA, suggesting that actimetric data could assist in developing objective methods for diagnosing eating disorders. Compared to traditional features, actimetric features provide more accurate classification of FA and SC, with daily activity analysis sufficient. The goal of future research should be to combine subjective and objective methods to enhance the analysis of actimetric data, to uncover hidden patterns of eating disorders. Future steps include evaluating FA diagnosis's sensitivity, specificity, and accuracy in a double-blind study, testing the methodology with a larger sample size, evaluating its effectiveness for small ethnic groups, and conducting longitudinal studies to assess its potential for mass screening. A growing number of wearable technologies and actigraphy are equipped with light sensors [87], allowing data analysis to be extended to circadian patterns of light exposure associated with metabolic disorders [88,89]. Circadian disruption can contribute to food addiction and eating disorders, such as binge eating and night eating [90].

In relation to the concept of a human bionic suite, IoT technology has the potential to improve monitoring and analysis capabilities by providing real-time data on motor activity. Real-time data processing and machine learning sensors could likely improve the accuracy and reliability of diagnoses, particularly when traditional methods fail. As a result of these promising results, wearable IoT devices can be effective in analyzing physiological indicators associated with FA and SC within digital health assistance, as well as supporting resilience to stress and mental health challenges within humans. To develop a more precise and personalized diagnostic approach for eating disorders, future research will focus on the integration of IoT and the refinement of machine learning algorithms.

**Author Contributions:**

Conceptualization, MiB, MaB, AV, DK, MM and DG; methodology, MiB, TT, LB, MaB, MM and AV; software, MaB, AV; validation, MaB, AV, TT, MM and DK; formal analysis, MiB, TT, MaB, and DG; investigation, MiB, TT, LB and DG; resources, MiB, DK, NRR, and DG; data curation, MiB, MaB and AV; writing—original draft preparation, MiB, and AV; writing—review and editing, TT, LB, MM , MaB, DK, NRR, and DG; visualization, MaB, and AV; supervision, MiB; project administration, MiB; funding acquisition, MiB, AV, NRR and DG; All authors have read and agreed to the published version of the manuscript.

**Acknowledgement:** This research is supported by Princess Nourah bint Abdulrahman University (PNU) Researchers Supporting Project number (PNURSP2025R194), Princess Nourah bint Abdulrahman University, Riyadh, Saudi Arabia.

**Funding:** This research (Sections 2.7, 2.8, 3.1-3.3, 5) was funded by the Russian Science Foundation (grant no. 22-11-00055-P, https://rscf.ru/en/project/22-11-00055/, accessed on 10 June 2025). This research (Sections 2.1-2.4) is also supported by West Siberian Science and Education Center, Government of Tyumen District, Decree of 20.11.2020, No. 928rp (Head: GD). The study (Sections 1, 2.5, 2.6, 3.4, 4) was conducted within the framework of a scientific project at the Institute of Physiology of the Federal Research Center of the Komi Scientific Center of the Ural Branch of the Russian Academy of Sciences FUUU-2022-0066 [No. 1021051201895-9].

**Institutional Review Board Statement:** The study adhered to the tenets of the Declaration of Helsinki and research program was approved by the Ethics Committee of the Institute of Physiology of the Komi Scientific Center of the Ural Branch of the Russian Academy of Sciences (Protocol # 3, 26 March 2019). Privacy rights of human subjects have been observed. Each participant signed an informed consent form for experimentation with human subjects.

**Informed Consent Statement:** Informed consent was obtained from all subjects involved in the study.

**Data Availability Statement:** The data are available from the authors upon reasonable request.

**Conflicts of Interest:** The authors declare no conflicts of interest.